\documentclass[10pt,twocolumn,letterpaper]{article}

\usepackage{wacv}
\usepackage{times}
\usepackage{epsfig}
\usepackage{graphicx}
\usepackage{color}
\usepackage{graphics} 
\usepackage{epsfig} 
\usepackage{amsmath} 
\usepackage{amssymb}  
\usepackage{epstopdf}
\usepackage{mathtools}
\usepackage{stfloats}
\usepackage{mathtools}
\usepackage{wrapfig}
\usepackage{enumerate}
\usepackage{booktabs}
\usepackage{makecell}
\usepackage{multirow}
\usepackage{amssymb}
\usepackage{pifont}

\newcommand{\PreserveBackslash}[1]{\let\temp=\\#1\let\\=\temp}
\newcommand{\tabincell}[2]{\begin{tabular}{@{}#1@{}}#2\end{tabular}}

\newcolumntype{C}[1]{>{\PreserveBackslash\centering}p{#1}}
\newcolumntype{R}[1]{>{\PreserveBackslash\raggedleft}p{#1}}
\newcolumntype{L}[1]{>{\PreserveBackslash\raggedright}p{#1}}

\usepackage{cite}
\usepackage[pagebackref=true,breaklinks=true,letterpaper=true,colorlinks,bookmarks=false]{hyperref}

\wacvfinalcopy 

\def\wacvPaperID{****} 

\ifwacvfinal\fi
\begin{document}

\title{Scene Parsing via Dense Recurrent Neural Networks with Attentional Selection}

\author{Heng Fan \;\;\; Peng Chu \;\;\; Longin Jan Latecki \;\;\; Haibin Ling \\
	Department of Computer and Information Sciences, Temple University, USA\\
	{\tt\small \{hengfan,tug29183,latecki,hbling\}@temple.edu}
}

\maketitle
\ifwacvfinal\thispagestyle{empty}\fi

\begin{abstract}
   Recurrent neural networks (RNNs) have shown the ability to improve scene parsing through capturing long-range dependencies among image units. In this paper, we propose dense RNNs for scene labeling by exploring various long-range semantic dependencies among image units. Different from existing RNN based approaches, our dense RNNs are able to capture richer contextual dependencies for each image unit by enabling immediate connections between each pair of image units, which significantly enhances their discriminative power. Besides, to select relevant dependencies and meanwhile to restrain irrelevant ones for each unit from dense connections, we introduce an attention model into dense RNNs. The attention model allows automatically assigning more importance to helpful dependencies while less weight to unconcerned dependencies. Integrating with convolutional neural networks (CNNs), we develop an end-to-end scene labeling system. Extensive experiments on three large-scale benchmarks demonstrate that the proposed approach can improve the baselines by large margins and outperform other state-of-the-art algorithms.
\end{abstract}

\section{Introduction}
\label{sec_intro}

Scene parsing or scene labeling, aiming to assign one of predefined labels to each pixel in an image, is usually formulated as a pixel-level classification problem. Inspired by the success of convolutional neural networks (CNNs) in image classification~\cite{krizhevsky2012imagenet,simonyan2014very,he2016deep}, CNNs have drawn increasing interests in scene labeling and demonstrated promising performance~\cite{farabet2013learning,long2015fully,noh2015learning,badrinarayanan2015segnet,girshick2014rich}. A potential issue, however, for CNN based methods is that only limited contextual cues from a local region (i.e., {\it receptive field}) in CNNs are explored for classification, which is prone to cause ambiguities for visually similar pixels of different categories. For example, the `sand' pixels can be visually indistinguishable from `road' pixels even for human with limited context. To alleviate this
\begin{figure}[!htb]
	\centering
	\includegraphics[width=0.97\linewidth]{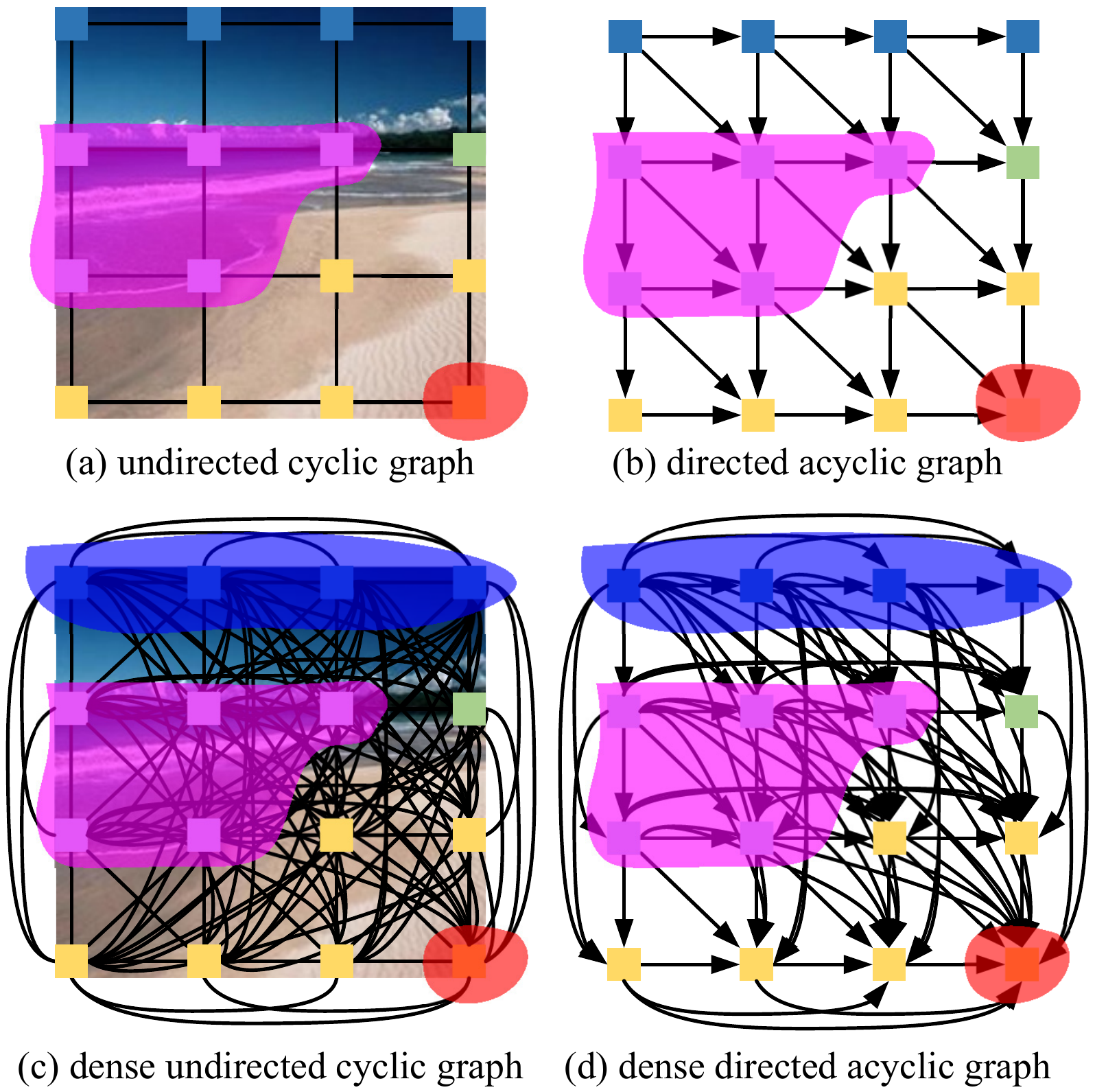}\\
	\caption{Image (a) shows the image of UCG structure as in~\cite{shuai2017scene}, and (b) demonstrates one of four DAG decompositions. Unlike~\cite{shuai2017scene}, we represent an image with dense UCG (D-UCG) as shown in (c), and (d) displays one of four dense DAGs (D-DAGs). Compared to plain UCG and DAG, our D-UCG and D-DAG capture richer dependencies in an image. Best viewed in color.}
	\label{fig:fig1}
\end{figure}
issue, a natural solution is to use rich context to discriminate locally ambiguous pixels~\cite{chen2016deeplab,yu2015multi,liu2015parsenet,zhao2017pyramid}. In these methods, nevertheless, the long-range contextual dependencies among different image regions are still not effectively explored, which are crucial in scene parsing.

Motivated by the ability of capturing long-range dependency among sequential data, recurrent neural networks (RNNs)~\cite{elman1990finding} have recently been employed to model semantic dependencies in images for scene labeling~\cite{byeon2015scene,li2016lstm,shuai2017scene,liang2016semantic,visin2016reseg,shuai2016dag,fan2017rgb}, allowing us to perform long-range inferences to discriminate ambiguous pixels.

To model the dependencies among image units, an effective way~\cite{shuai2017scene,zuo2016learning} is to represent the image with an undirected cyclic graph (UCG) in which the image units are vertices and their interactions are encoded by undirected edges (see Fig.~\ref{fig:fig1}(a)). Due to the loopy structure of UCG, however, it is hard to directly apply RNNs to model dependencies in an image. To handle this problem, a UCG is approximated with several directed acyclic graphs (DAGs) (see Fig.~\ref{fig:fig1}(b)). Then several DAG-structured RNNs are adopted to model the dependencies in these DAGs.

\subsection{Motivation}

Though these DAG-structured RNNs can model dependencies in images, some useful information may be discarded. For instance in Fig.~\ref{fig:fig1}(a), to correctly distinguish a `sand' unit (marked in red region) from a `road' one, DAG-structured RNNs can use the dependencies of `water' units (marked in pink region) from its adjacent neighbors. However, the `water' information may be decaying because it needs to pass through conductors (i.e., the adjacent neighbors of this `sand' unit). Instead, a better way is to directly use dependencies from `water' units to recognize the `sand' unit. To such end, we propose dense RNNs to fully explore abundant dependencies in images for scene parsing.

Analogous to CNNs, DAG-structured RNNs can be unfolded to a feed-forward network where each vertex is a layer and a directed edge between two layers represents information flow (i.e., dependency relationship between two vertexes). The dependency information in an image flows from the first layer (i.e., the start vertex at top-left corner in Fig.~\ref{fig:fig1}(b)) to the last layer (i.e., the end vertex at bottom-right corner in Fig.~\ref{fig:fig1}(b)). Inspired by the superior performance of recently proposed DenseNet~\cite{huang2016densely} in image recognition, which introduces dense connections among layers to improve information flow in CNNs, we propose to add more connections into the RNN feed-forward network as well (see Fig.~\ref{fig:fig1}(d)), to incorporate richer dependency information among image units.

Despite abundant dependencies from dense connections, we argue that not all dependencies are equally helpful to recognize a specific image region. For example in Fig.~\ref{fig:fig1}(d), the `sky' units in blue region are not useful to distinguish the `sand' unit in the red region from the `road' unit. In contrast, the dependencies from `water' units in the pink region are more crucial to infer its label. Therefore, more importance should be assigned to the dependencies from `water' units, which motivates us to integrate an attention model into dense RNNs to select more useful dependencies.

\subsection{Contribution}

The {\bf first contribution} of this work is the dense RNNs, which capture richer dependencies for image units from various abundant connections. Unlike previous approaches representing an image as a UCG, we formulate each image with a dense UCG (D-UCG), which is a complete graph. In D-UCG, each pair of vertexes are connected with an undirected edge (see Fig.~\ref{fig:fig1}(c)). By decomposing the D-UCG into several dense DAGs (D-DAGs), we propose the DAG-structured dense RNNs (DD-RNNs) to model dependencies in an image (see Fig.~\ref{fig:fig1}(d)). Compared with plain DAG-structured RNNs, our DD-RNNs can gain richer dependencies from various levels. For instance in Fig.~\ref{fig:fig1}(c), to correctly recognize the `sand' unit in the red region, in addition to the dependencies from its neighbors, DD-RNNs enable the {\it firsthand} use of dependencies from `water' units in the pink region to improve its discriminability.

Although DD-RNNs are capable of capturing vast dependencies through dense connections, for a specific image unit, certain dependencies are {\it irrelevant} to help improve discriminative power. To tackle this issue, we make the {\bf second contribution} by introducing a novel attention model into DD-RNNs. The attention model is able to automatically select {\it relevant} and meanwhile restrain {\it irrelevant} dependency information for image units, further enhancing their discriminative power.

Last but not least, the {\bf third contribution} is to implement an end-to-end labeling system based on our DD-RNNs. For validation, we test our method on three benchmarks: PASCAL Context~\cite{mottaghi2014role}, MIT ADE20K~\cite{zhou2017scene} and Cityscapes~\cite{cordts2016cityscapes}. In these experiments the proposed approach significantly improves the baselines and outperforms other state-of-the-art methods.

\section{Related Work}
\label{sec_rel}

\noindent
{\bf Scene parsing.} Scene parsing has drawn extensive attentions in recent decades. Early efforts mainly focus on the graphical model with hand-crafted features~\cite{liu2011sift,tighe2013finding,yang2014context,gould2009decomposing}. Despite great progress, these methods are restricted due to the use of hand-crafted features.

Inspired by the success in image recognition~\cite{krizhevsky2012imagenet,simonyan2014very,he2016deep}, CNNs have been extensively explored for scene parsing. Long {\it et al.}~\cite{long2015fully} propose a scene labeling method by transforming standard CNNs for classification into fully convolutional networks (FCN), resulting in significant performance gains. To generate desired full-resolution predictions, various methods are proposed to upsample low-resolution feature maps to high-resolution feature maps for final prediction~\cite{noh2015learning,badrinarayanan2015segnet,lin2016refinenet}. In order to remit boundary problem in predictions, graphical models such as Conditional Random Field (CRF) or Markov Random Field (MRF) are introduced into CNNs~\cite{chen2016deeplab,zheng2015conditional,liu2015semantic}. As a pixel-level classification problem, contexts are crucial role in scene labeling to distinguish visually similar pixels of different categories. The work of~\cite{yu2015multi} introduces the dilated convolution into CNNs to aggregate multi-scale context. Liu {\it et al.}~\cite{liu2015parsenet} suggest an additional branch in CNNs to incorporate global context for scene parsing. In~\cite{zhao2017pyramid}, Zhao {\it et al.} propose a spatial pyramid pooling module to fuse contexts from different levels, showing superior performance in scene parsing. Zhang {\it et al.}~\cite{zhang2018context} introduce an context encoding module into CNNs to improve parsing performance.

\vspace{0.3em}
\noindent
{\bf RNNs on computer vision.} With the capability of modeling spatial dependencies in images, RNNs~\cite{elman1990finding} have been applied to many computer vision tasks such as image completion~\cite{oord2016pixel}, handwriting recognition~\cite{graves2009offline}, image classification~\cite{zuo2016learning}, visual tracking~\cite{fan2017sanet}, skin detection~\cite{zuo2017combining} and so forth. Considering the importance of spatial dependencies in an image to distinguish ambiguous pixels, there are attempts to applying RNNs for scene labeling.

The work of~\cite{byeon2015scene} explores the two-dimensional long-short term memory (LSTM) networks for scene parsing by taking into account the spatial dependencies of pixels in images. Stollenga {\it et al.}~\cite{stollenga2015parallel} introduce a parallel multi-dimensional LSTM for image segmentation. Liang {\it et al.}~\cite{liang2016semantic} propose a graph based LSTM to model the dependencies among different superpixels. The work of~\cite{liang2016semantic} applies a local-global LSTM model on object parsing. Visin {\it et al.}~\cite{visin2016reseg} suggest to utilize multiple linearly structured RNNs to model horizontal and vertical dependencies among image units for scene labeling. Li {\it et al.}~\cite{li2016lstm} extend this method by substituting RNNs with LSTM and apply it to RGB-D scene labeling. Qi~\cite{qi2016hierarchically} proposes the gated recurrent units (GRUs) to model long-range context. Especially, to exploit more spatial dependencies in images, Shuai {\it et al.}~\cite{shuai2017scene} propose to represent an image with a UCG. By decomposing UCG into several DAGs, they then propose to use DAG-structured RNNs to model dependencies among image units.

\vspace{0.3em}
\noindent
{\bf Attention model.} The attention model, being successfully applied in Natural Language Processing (NLP) such as machine translation~\cite{bahdanau2014neural} and sentence summarization~\cite{rush2015neural}, has drawn increasing interest in computer vision. Xu {\it et al.}~\cite{xu2015show} propose to leverage an attention model to find out regions of interest in images which are relevant in generating next word. In~\cite{chen2016attention}, Chen {\it et al.} propose a scale attention model for semantic segmentation by adaptively merging outputs from different scales. In~\cite{abdulnabi2017episodic}, the attention model is utilized to assign importance to different regions for context modeling in images. The work of~\cite{lu2016hierarchical} introduces a co-attention model to combine question and image features for question answering. Chu {\it et al.}~\cite{chu2017multi} utilize attention model to fuse multi-context for human pose estimation.

\vspace{0.3em}
\noindent
{\bf Our approach.} In this paper, we focus on how to effectively exploit abundant dependencies in images and introduce the dense RNNs module. Our approach is related to but different from previous RNN approaches (e.g., DAG-structured RNNs~\cite{shuai2017scene} and linearly structured RNNs~\cite{visin2016reseg} or LSTM~\cite{li2016lstm}), in which each image unit only receives dependency information from its limited neighbors and considerable useful dependencies are thrown away. In contrast, we propose to add dense paths into RNNs to enable immediate long-range dependencies. Consequently, each image unit can directly `see' dependencies in the whole image, leading to more discriminative representation. It is worth noting that the idea of dense connections can not only used for graphical RNNs~\cite{shuai2017scene} but also easily applied to other linearly structured RNNs~\cite{visin2016reseg,li2016lstm}.

Furthermore, we introduce an attention model into dense RNNs. To the best of our knowledge, this work is the first to use attention mechanism in RNNs for scene parsing. Our attention model automatically selects relevant and restrains irrelevant dependencies for image units from dense connections, further improving their discriminabilities.

\section{The Proposed Approach}
\label{drnns}


\subsection{Review of DAG-structured RNNs}
\label{sec30}

The linear RNNs in~\cite{elman1990finding} are designated to deal with sequential data related tasks. Specifically, a hidden unit $h_t$ in RNNs at time step $t$ is represented with a non-linear function over current input $x_t$ and hidden layer at previous time step $h_{t-1}$, and the output $y_t$ is connected to the hidden unit $h_t$. Given an input sequence $\{x_t\}_{t=1,2,\cdots,T}$, the hidden unit and output at time step $t$ can be computed with
\begin{align}
	h_t = &\phi(Ux_t+Wh_{t-1}+b) \label{eq1} \\
	y_t = &\sigma(Vh_t+c)
\end{align}
where $U$, $V$ and $W$ represent transformation matrices, $b$ and $c$ are bias terms, and $\phi(\cdot)$ and $\sigma(\cdot)$ are non-linear functions, respectively. Since the inputs $\{x_t\}_{t=1,2,\cdots,T}$ are progressively stored in the hidden layers as in Eq. (\ref{eq1}), RNNs are able to preserve the memory of entire sequence and thus capture the long-range contextual dependencies.

For an image, the interactions among image units can be formulated as a graph in which the dependencies are forwarded through edges. The solution in~\cite{shuai2017scene} utilizes a standard UCG to represent an image (see again Fig.~\ref{fig:fig1}(a)). To break the loopy structure of UCG, \cite{shuai2017scene} further proposes to decompose the UCG into four DAGs along different directions (see Fig.~\ref{fig:fig1}(b) for a southeast example).

Let $\mathcal{G}=\{\mathcal{V}, \mathcal{E}\}$ denote the DAG as shown in Fig.~\ref{fig:fig1}(b), where $\mathcal{V}=\{v_{i}\}_{i=1}^N$ represents the vertex set of $N$ vertexes, $\mathcal{E}=\{e_{ij}\}_{i,j=1}^N$ represents the edge set, and $e_{ij}$ indicates a directed edge from $v_i$ to $v_j$. A DAG-structured RNN resembles the identical topology of $\mathcal{G}$, with a forward pass formulated as traversing $\mathcal{G}$ from  start vertex. In such modeling, the hidden layer of each vertex relies the hidden units of its adjacent predecessors (see Fig.~\ref{fig:fig2}(b)). For vertex $v_i$, its hidden layer $h_{v_i}$ and output $y_{v_i}$ are computed with
\begin{align}
	h_{v_i} = &\phi(Ux_{v_i}+W~~\sum\limits_{\mathclap{v_j\in{\mathcal{P}_{\mathcal{G}}(v_i)}}}{h_{v_j}}+b) \label{eq4} \\
	y_{v_i} = &\sigma(Vh_{v_i}+c)
\end{align}
where $x_{v_i}$ denotes the local feature at vertex $v_i$ and $\mathcal{P}_{\mathcal{G}}(v_i)$ represents the predecessor set of $v_i$ in $\mathcal{G}$. By storing local inputs into hidden layers and progressive forwarding among them with Eq. (\ref{eq4}), the discriminative power of each image unit is improved with dependencies from other units.

\subsection{Dense RNNs}
\label{sec31}

In DAG-structured RNNs, each image unit receives the dependencies from other units through recurrently forwarding information between adjacent units. Nevertheless, the useful dependency information may be potentially degraded after going through many conductors, resulting in a {\it dependency decaying} problem. For instance in Fig.~\ref{fig:fig1}(b), the most useful contextual cues from `water' units have to pass through conductors to arrive at the `sand' unit covered in the red region. A natural solution to remedy the problem of dependency decaying is to add additional paths between hidden layers of distant units and current image unit.

Inspired by the recently proposed DenseNet~\cite{huang2016densely} that introduces dense connections into CNNs, we propose DAG-structured dense RNNs (DD-RNNs) to model richer dependencies in an image. We first view a DAG-structured RNNs as unfolded to get a feed-forward network, where the dependency information in an image flows from start to end vertexes. Then, to capture richer dependencies in images (e.g., forthright dependencies among {\it non-adjacent} units in Fig.~\ref{fig:fig1}(b)), we introduce more connections in the RNN feed-forward network, resulting in the proposed DD-RNNs.

To achieve dense connections, we represent each image with a dense UCG (D-UCG), which is equivalent to a complete graph (see Fig.~\ref{fig:fig1}(c) for illustration). Compared to standard UCG, D-UCG allows each image unit to connect with all of other units. Because of the loopy property of D-UCG, we adopt the strategy as in~\cite{shuai2017scene} to decompose the D-UCG to four D-DAGs along four directions. One of the four D-DAGs along the southeast direction is shown in Fig.~\ref{fig:fig1}(d).

\begin{figure}[!t]
	\centering
	\includegraphics[width=0.97\linewidth]{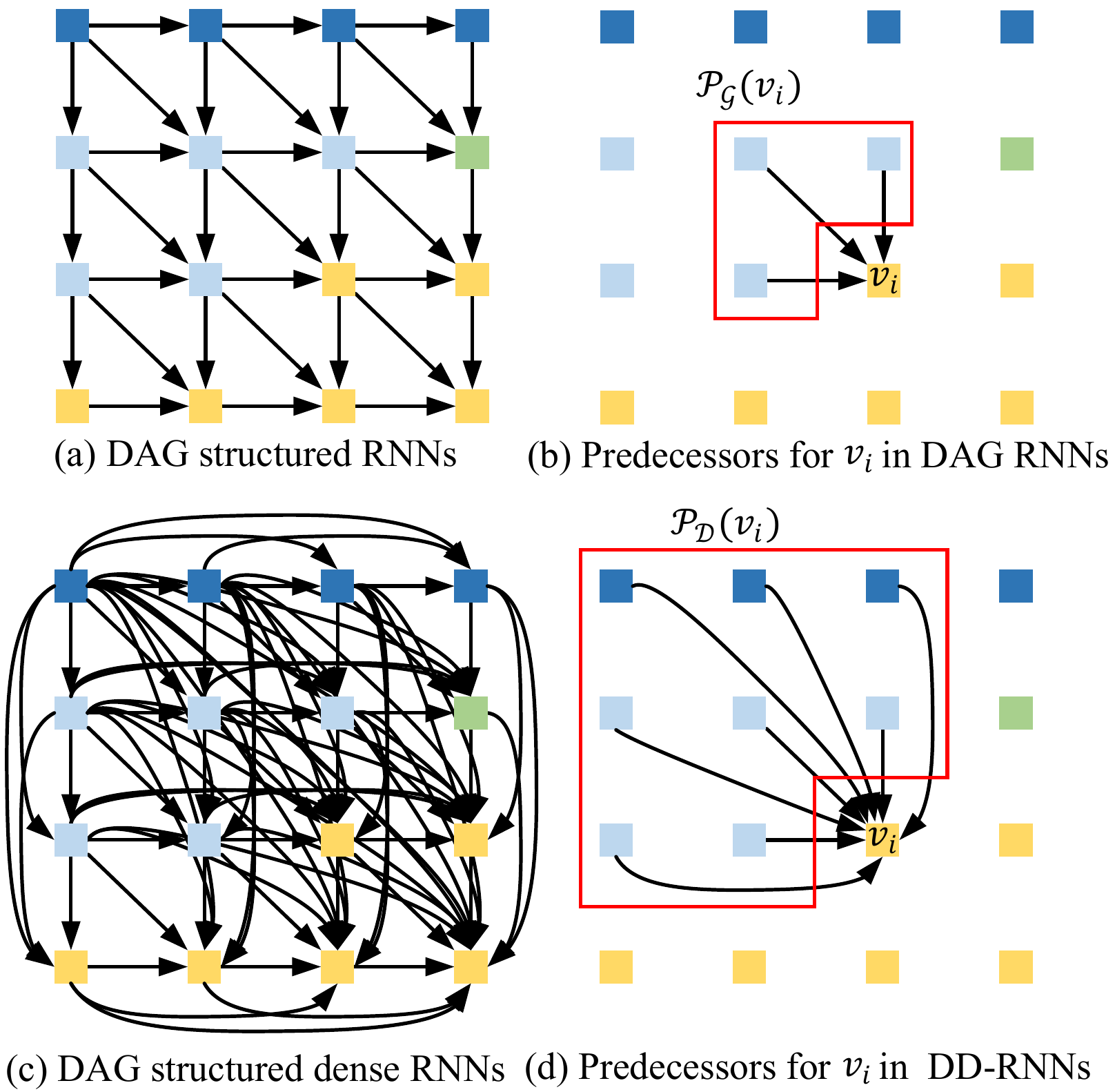}\\
	\caption{The illustration of difference between DAG-structured RNNs~\cite{shuai2017scene} and our DD-RNNs. Image (a) shows the DAG-structured RNNs along southeast direction, and in (b) the hidden layer of vertex $v_i$ relies on its three adjacent predecessors (see the red region in (b)). Image (c) is our DD-RNNs, and in (d) the hidden layer of $v_i$ is dependent on all its adjacent and non-adjacent predecessors (see the red region in (d)). Best viewed in color.}
	\label{fig:fig2}
\end{figure}

Let $\mathcal{D}$ represent the D-DAG in Fig.~\ref{fig:fig1}(d). The structure of DD-RNNs resembles the identical topology of $\mathcal{D}$ as in Fig. \ref{fig:fig2}(c). In DD-RNNs, the hidden layer of each vertex relies on the hidden units of all its \emph{adjacent} and \emph{non-adjacent} predecessors, which fundamentally differs from~\cite{shuai2017scene} in which the hidden unit of each vertex only relies on hidden units of its adjacent predecessors (see Fig.~\ref{fig:fig2}(b)). The forward pass at the vertex $v_i$ in DD-RNNs is expressed as
\begin{align}
	\hat{h}_{v_i} = &\sum\limits_{\mathclap{v_j\in{\mathcal{P}_{\mathcal{D}}(v_i)}}}{h_{v_j}} \label{eq33}\\
	h_{v_i} = &\phi(Ux_{v_i}+W\hat{h}_{v_i}+b) \label{eq44} \\
	y_{v_i} = &\sigma(Vh_{v_i}+c) \label{eq55}
\end{align}
where $\mathcal{P}_{\mathcal{D}}(v_i)$ is the dense predecessor set of $v_i$ in D-DAG $\mathcal{D}$, and it contains both adjacent and non-adjacent predecessors (see Fig.~\ref{fig:fig2}(d)). Compared to the DAG-structured RNNs in~\cite{shuai2017scene}, our DD-RNNs are able to capture richer dependencies in an image through various dense connections.

A concern arisen naturally from the dense model is the complexity. In fact, it is unrealistic to directly apply DD-RNN to pixels of an image. Fortunately, neither is it necessary. As described in Section~\ref{sec33}, we apply DD-RNN to a high layer output of existing CNN models. Such strategy largely reduces the computational burden -- as summarized in Table 7, our final system runs faster than many state-of-the-arts while achieving better labeling accuracies.

\subsection{Attention model in DD-RNNs}
\label{sec32}

\begin{figure*}[!htb]
	\centering
	\begin{tabular}{c}
		\includegraphics[width=0.97\linewidth]{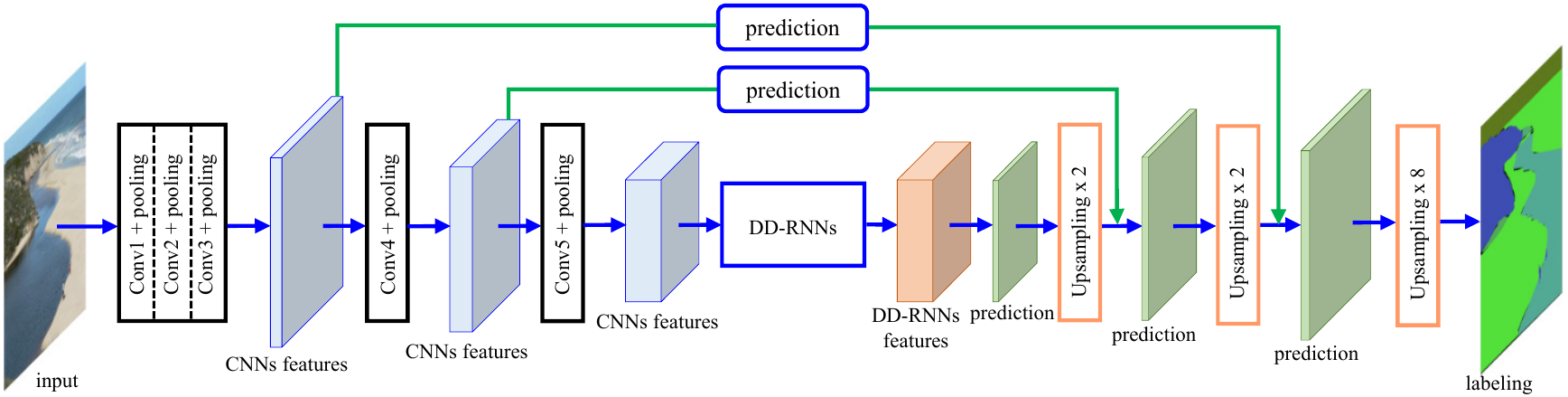}\\
	\end{tabular}
	\caption{The architecture of our full system. The DD-RNNs are placed on the top of feature maps obtained from the last convolutional block to model long-range dependencies in an image, and the deconvolution is used to upsample the predictions. Low-level and high-level features are combined through skip strategy for final labeling (see the green arrows). Best viewed in color.}
	\label{fig:fig3}
\end{figure*}

For the hidden layer at vertex $v_i$, it receives various dependency information from predecessors through dense connections. However, the dependencies from different predecessors are not always equally helpful to improve the discriminative representation (see Fig.~\ref{fig:fig2}(d)). For example, to distinguish the `sand' units from visually alike `road' units in a beach scene, the most important contextual cues are probably the dependencies from `water' units instead of other units such as `sky' or `tree'. In this case, we term the relation from `water' units as relevant dependencies while the information from `sky' or `tree' units as irrelevant ones.

To encourage relevant dependencies and meanwhile restrain irrelevant ones for each image unit, we introduce a soft attention model~\cite{bahdanau2014neural} into DD-RNNs. In~\cite{bahdanau2014neural}, the attention model is employed to softly assign importance to input words in a sentence when predicting a target word for machine translation. In this paper, we leverage attention model to select more relevant and useful dependencies for each image unit. To this end, we do not directly use Eq. (\ref{eq33}) and (\ref{eq44}) to model the relationships between $h_{v_i}$ and its predecessors. Instead, we employ the following expression to model the dependency between $h_{v_i}$ and one of its predecessors $h_{v_j}$
\begin{equation}\label{eq6}
	h_{v_i,v_j} = \phi(Ux_{v_i}+Wh_{v_j}+b)
\end{equation}
where $h_{v_j}$ represents the hidden layer of a predecessor $v_j\in\mathcal{P}_{\mathcal{D}}(v_i)$ of $v_i$. The $h_{v_i,v_j}$ in Eq. (\ref{eq6}) models dependency information from $h_{v_j}$ for $h_{v_i}$. The final hidden unit $h_{v_i}$ at $v_i$ is obtained by summarizing all $h_{v_i,v_j}$ with attentional weights, as computed by
\begin{equation}\label{eq7}
	h_{v_i} = \sum_{\mathclap{v_j\in{\mathcal{P}_{\mathcal{D}}(v_i)}}}{h_{v_i,v_j}w_{v_i,v_j}}
\end{equation}
where the attention weight $w_{v_i,v_j}$ for $h_{v_j}$ reflects the relevance of the predecessor $v_j$ to $v_i$, calculated by
\begin{equation}\label{eq8}
	w_{v_i,v_j} = \frac{\mathrm{exp}(z^{\mathrm{T}}h_{v_i,v_j})}{\sum\limits_{\mathclap{v_k\in{\mathcal{P}_{\mathcal{D}}(v_i)}}} ~\exp(z^{\mathrm{T}}h_{v_i,v_k})}
\end{equation}
where $z^{\mathrm{T}}$ represents a transformation matrix.

With the above attention model, we replace Eq. (\ref{eq33}) and (\ref{eq44}) with Eq. (\ref{eq6}) and (\ref{eq7}) for a forward pass at $v_i$ in DD-RNNs. By using stochastic gradient descent (SGD) method, the attentional DD-RNNs can be trained in an end-to-end manner.

\subsection{Full labeling system}
\label{sec33}

Before showing the full labeling system, we first introduce the decomposition of D-UCG. As in~\cite{shuai2017scene}, we decompose the D-UCG $\mathcal{U}$ into a set of D-DAGs represented with $\{\mathcal{D}^{l}\}_{l=1}^L$, where $L$ is the number of D-DAGs. Since Equation (\ref{eq7}) only computes the hidden layer at vertex $v_i$ in one of $L$ D-DAGs, the final output $\hat{y}_{v_i}$ at $v_i$ is derived by aggregating the hidden layers at $v_i$ from all D-DAGs. The mathematical formulation for this process is expressed as
\begin{align}
	h_{v_i,v_j}^{l} = &\phi(U^{l}x_{v_i}+W^{l}h_{v_j}^{l}+b_l) \label{eq12} \\
	h_{v_i}^{l} = &\sum\limits_{\mathclap{v_j\in{\mathcal{P}_{\mathcal{D}^{l}}(v_i)}}}{h_{v_i,v_j}^{l}w_{v_i,v_j}^{l}} \label{eq13}\\
	\hat{y}_{v_i} = &\sigma(\sum\nolimits_{l=1}^{L}V^{l}h_{v_i}^{l}+c) \label{eq14}
\end{align}
With the equations above, the proposed DD-RNNs can be used to capture abundant dependencies among image units.

We develop an end-to-end scene labeling system by integrating our approach with CNNs for scene parsing as shown in Fig.~\ref{fig:fig3}. The proposed DD-RNNs are placed on the top of feature maps obtained after the last convolutional block to model long-range dependencies in the input image, and the deconvolution operations are used to upsample the predictions. To produce the desired input size of labeling result, we utilize the deconvolution~\cite{zeiler2011adaptive} to upsample predictions. Taking into account both spatial and semantic information for scene labeling, we adopt the skip strategy~\cite{long2015fully} to combine low-level and high-level features. The whole system is trained end-to-end with the pixel-wise cross-entropy loss.

\begin{table*}[htbp]
	\centering
	\caption{Baseline comparisons of mIoU (\%) with different backbones on PASCAL Context~\cite{mottaghi2014role}, MIT ADE20K (validation set)~\cite{zhou2017scene} and Cityscapes (validation set)~\cite{cordts2016cityscapes}.}
	\begin{tabular}{L{2.5cm}C{1.7cm}C{1.7cm}C{1.7cm}C{1.7cm}C{1.7cm}C{1.7cm}}
		\hline
		& \multicolumn{2}{c}{PASCAL Context~\cite{mottaghi2014role}} & \multicolumn{2}{c}{MIT ADE20K~\cite{zhou2017scene}} & \multicolumn{2}{c}{Cityscapes~\cite{cordts2016cityscapes}}  \\
		\cline{2-7}
		& VGG-16 & ResNet-101 & VGG-16 & ResNet-101 & VGG-16 & ResNet-101  \\
		\hline
		Baseline FCN & 35.6  & 40.3  & 28.7  & 35.1  & 64.7  & 68.9   \\
		FCN+CRF & 40.1  & 43.8  & 30.8  & 36.2  & 66.7  &  69.2  \\
		FCN+DAG-RNN & 41.3  & 45.1  & 32.1  & 37.4  & 70.2  & 75.5  \\
		FCN+DD-RNN & {\bf 44.9}  & {\bf 49.3}  & {\bf 35.7}  & {\bf 40.9}  & {\bf 72.3}  & {\bf 78.2}   \\
		\hline
	\end{tabular}%
	\label{tab:baseline}%
\end{table*}%

\section{Experimental Results}
\label{sec_res}

\vspace{0.3em}
\noindent
{\bf Implementation details.} In order validate the effectiveness of the proposed DD-RNNs, we develop two labeling systems by integrating our DD-RNNs with two different architectures: the VGG-16~\cite{simonyan2014very} and the ResNet-101~\cite{he2016deep}. The DD-RNNs are employed to model dependencies among image units in output of the last convolutional block (Fig.~\ref{fig:fig3}). The network takes $512\times{512}$ images as inputs, and outputs the labeling results with the same resolution. When evaluating, the labeling results are resized to the original input size. The dimension of input, hidden and output units for DD-RNNs is set to 512. The two non-linear activations $\phi$ and $\sigma$ are {\it ReLU} and {\it softmax} functions, respectively. The full networks are end-to-end trained with standard SGD method. For convolutional blocks, the learning rate is initialized to be $10^{-4}$ and decays exponentially with the rate of 0.9 after 10 epochs. For D-RNNs, the learning rate is initialized to be $10^{-2}$ and decays exponentially with the rate of 0.9 after 10 epochs. The batch sizes for both training and testing phases are set to 1. The results are reported after 50 training epochs. The networks are implemented in Matlab using MatConvNet~\cite{vedaldi2015matconvnet} on a single Nvidia GeForce TITAN GPU with 12GB memory.

\vspace{0.3em}
\noindent
{\bf Datasets.} We test our method on the large-scale PASCAL Context~\cite{mottaghi2014role}, MIT ADE20K~\cite{zhou2017scene} and Cityscapes~\cite{cordts2016cityscapes}.

The PASCAL Context contains 10,103 images annotated into 540 classes, where 4,998 images are used for training and the rest for testing. Similar to other literatures, we only consider the most frequent 59 classes for evaluation.

The recent MIT ADE20K consists of 20,000 images in training set and 2,000 images in validation set. There are total 150 semantics classes in the dataset.

The Cityscapes contains 5000 images of street traffic scene, where 2975 images are used for training, 500 images for validation, and the rest for testing. In total, 19 classes are considered for training and evaluation.

\vspace{0.3em}
\noindent
{\bf Evaluation metrics.} As in~\cite{long2015fully}, we utilize mean Intersection over Union (mIoU\%) for evaluation.

\subsection{Baseline comparisons}

To better analyze our method, we develop several baselines to prove its effectiveness:

{\bf Baseline FCN} is implemented by removing our attentional DD-RNNs from networks. Note that the baseline FCN differs from FCN-8s~\cite{long2015fully} because we discard two fully
connected layers. Other settings remain the same as in FCN-8s~\cite{long2015fully}.

{\bf FCN+CRF} is implemented by applying CRF~\cite{krahenbuhl2011efficient} to perform post-processing on the results of baseline FCN.

{\bf FCN+DAG-RNN} is implemented by substituting the attentional DD-RNNs with {\it plain} DAG-RNN. Note that FCN+DAG-RNN varies from~\cite{shuai2017scene} because we do not use class weighting strategy and larger conventional kernel in our labeling system.

{\bf FCN+DD-RNNs} represents the proposed scene labeling method.

Table~\ref{tab:baseline} shows the quantitative results between different baselines and our approach with two backbones. All CRF, DAG-RNN and DD-RNNs can improve the performance of baseline FCN. More specific, our method obtains mIoU gains of 9.3\%, 7.0\% and 7.6\% with VGG-16 and of 9.0\%, 5.8\% and 9.3\% with ResNet-101 on three datasets, and outperforms other two baselines using CRF and DAG-RNN.

\subsection{Comparison results on PASCAL Context}

\begin{table}[htbp]
	\centering
	\caption{Quantitative comparisons on PASCAL Context~\cite{mottaghi2014role} (59 classes).}
	\begin{tabular}{r|ccc}
		\hline
		Algorithm & Backbone   & mIoU (\%) \\
		\hline
		\hline
		CAMN~\cite{abdulnabi2017episodic}    & VGG-16       & 41.2 \\
		PixelNet~\cite{bansal2017pixelnet} & VGG-16        & 41.4 \\
		FCN-8s~\cite{long2015fully}  & VGG-16        & 38.2 \\
		HO-CRF~\cite{arnab2016higher}  & VGG-16         & 41.3 \\
		BoxSup~\cite{dai2015boxsup}  &  VGG-16         & 40.5 \\
		ParseNet~\cite{liu2015parsenet} & VGG-16         & 40.4 \\
		ConvPP-8~\cite{xie2016top} & VGG-16         & 41.0 \\
		CNN-CRF~\cite{lin2016efficient} & VGG-16       & 43.3 \\
		CRF-RNN~\cite{zheng2015conditional} & VGG-16         & 39.3 \\
		DAG-RNN~\cite{shuai2017scene} & VGG-16      & 42.6 \\
		DAG-RNN-CRF~\cite{shuai2017scene} & VGG-16       & 43.7 \\
		DeepLab v2-CRF~\cite{chen2016deeplab1} &ResNet-101        & 44.4 \\
		GCE~\cite{hung2017scene} &ResNet-101        & 46.5 \\
		RefineNet~\cite{lin2016refinenet} &ResNet-101        & 47.1 \\
		\hline
		\hline
		DD-RNNs   &  VGG-16       &  44.9 \\
		DD-RNNs   & ResNet-101      & {\bf 49.3} \\
		\hline
	\end{tabular}
	\label{tab:tab1}
\end{table}

The quantitative comparisons to state-of-the-art methods are summarized in Table~\ref{tab:tab1}. Benefiting from deep CNNs, FCN-8s~\cite{long2015fully} achieves promising result with mIoU of 38.2\%. In order to alleviate boundary issue in FCN-8s, CRF-RNN~\cite{zheng2015conditional} and DeepLab v2-CRF~\cite{chen2016deeplab1} use probabilistic graphical model such as CRF in CNNs, and obtain mIoUs of 39.3\% and 39.6\%, respectively. Other approaches such as CAMN~\cite{abdulnabi2017episodic},  ParseNet~\cite{liu2015parsenet} and GCE~\cite{hung2017scene} suggest to improve performance by incorporating global contextual information and obtain mIoUs of 41.2\%, 40.4\% and 46.5\%. Despite improvements, these methods ignore long-range dependencies in
\begin{figure}[!htb]
	\centering
	\begin{tabular}{@{}C{1.65cm}@{}C{1.65cm}@{}C{1.65cm}@{}C{1.65cm}@{}C{1.65cm}@{}}
		\includegraphics[width=1.6cm]{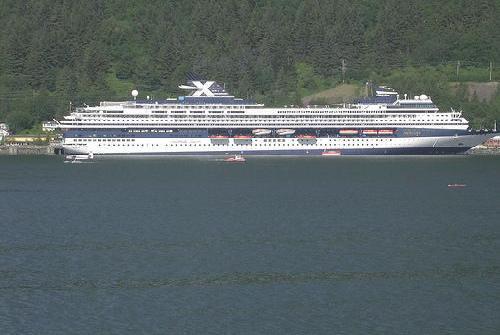}&\includegraphics[width=1.6cm]{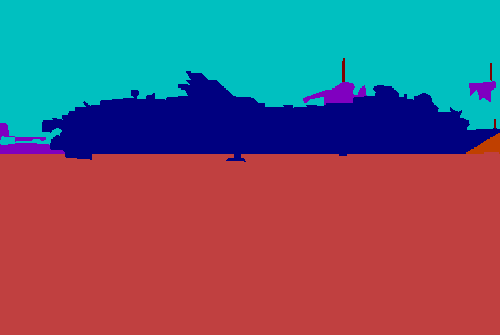}& \includegraphics[width=1.6cm]{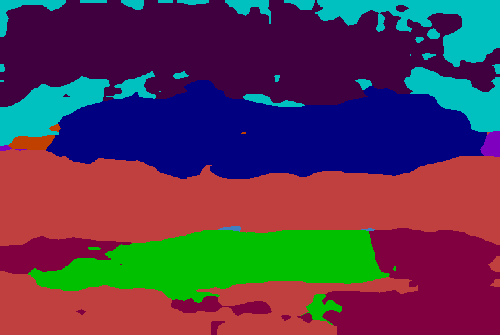}&\includegraphics[width=1.6cm]{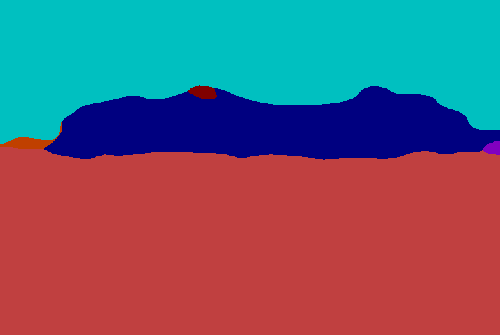}&\includegraphics[width=1.6cm]{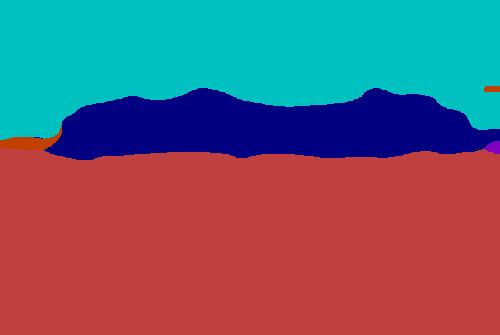}\\
		\includegraphics[width=1.6cm]{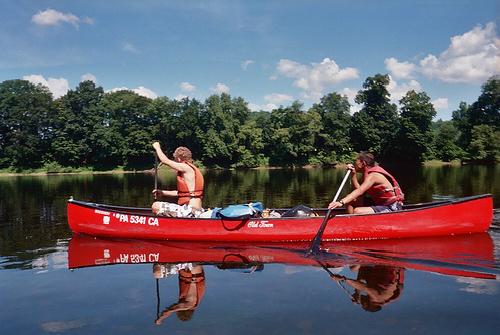}&\includegraphics[width=1.6cm]{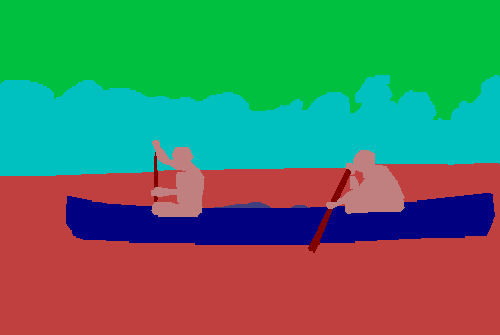}& \includegraphics[width=1.6cm]{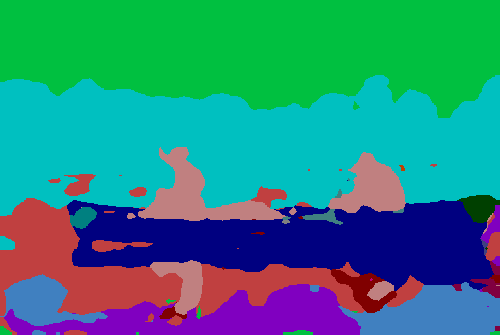}&\includegraphics[width=1.6cm]{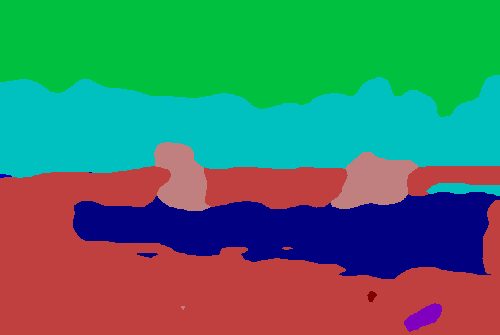}&\includegraphics[width=1.6cm]{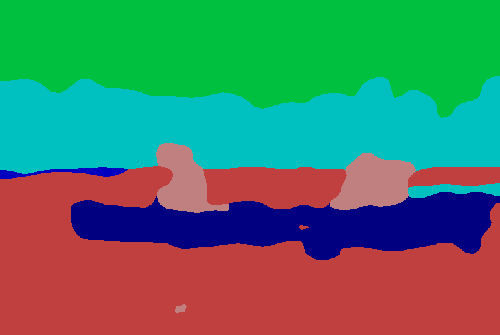}\\
		\includegraphics[width=1.6cm]{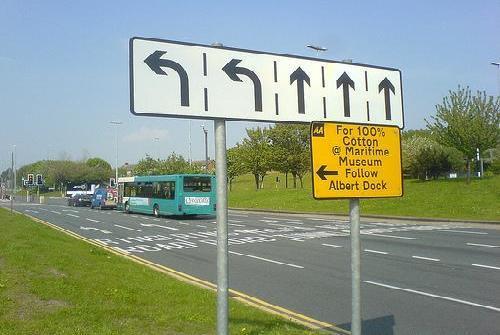}&\includegraphics[width=1.6cm]{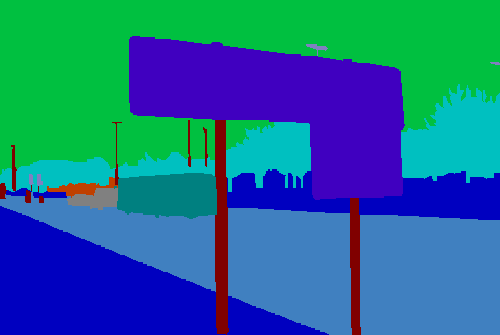}& \includegraphics[width=1.6cm]{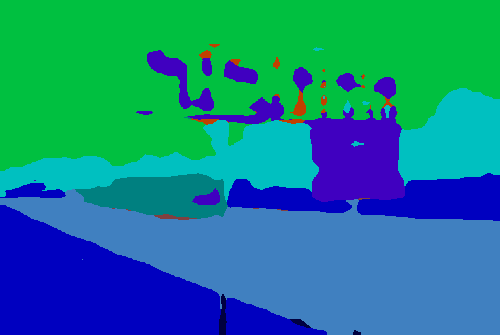}&\includegraphics[width=1.6cm]{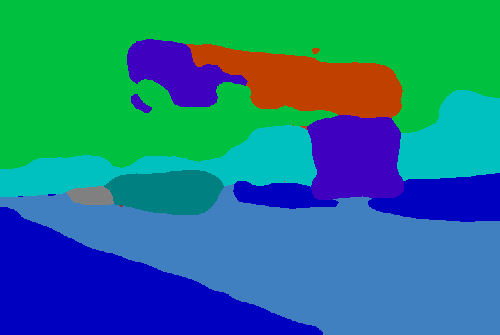}&\includegraphics[width=1.6cm]{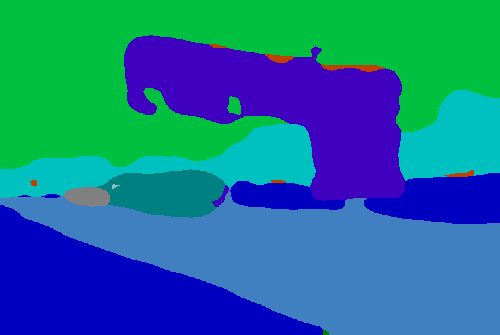}\\
		\includegraphics[width=1.6cm]{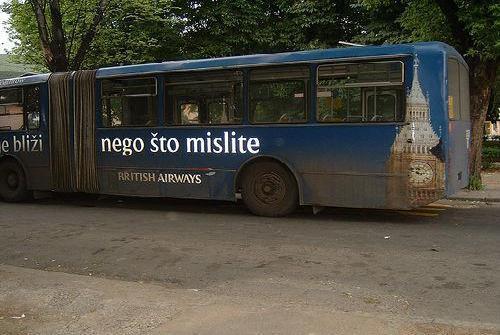}&\includegraphics[width=1.6cm]{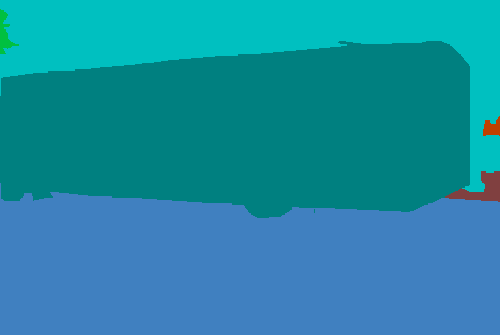}& \includegraphics[width=1.6cm]{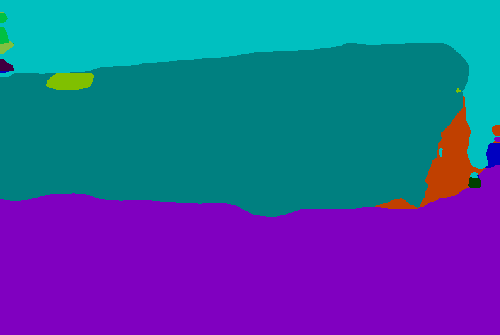}&\includegraphics[width=1.6cm]{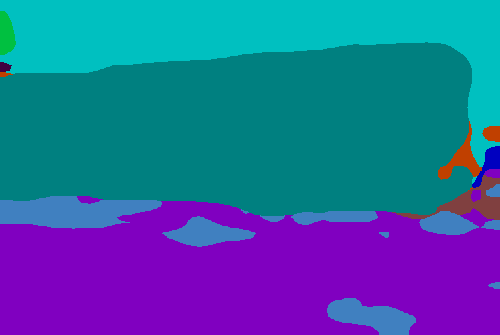}&\includegraphics[width=1.6cm]{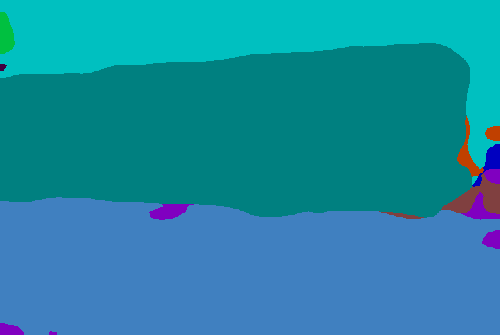}\\
		\includegraphics[width=1.6cm]{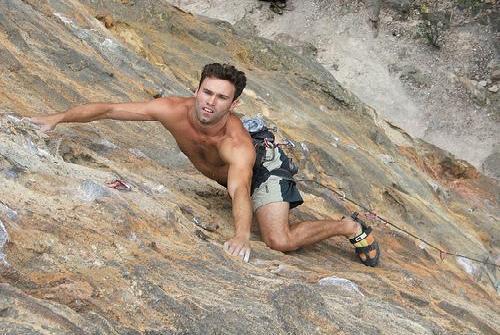}&\includegraphics[width=1.6cm]{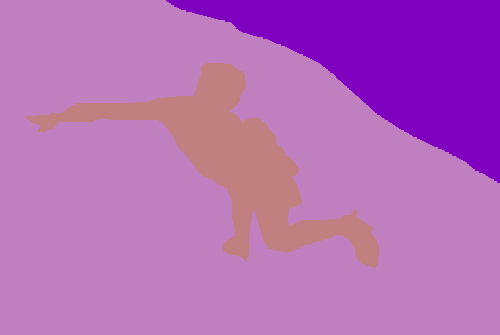}& \includegraphics[width=1.6cm]{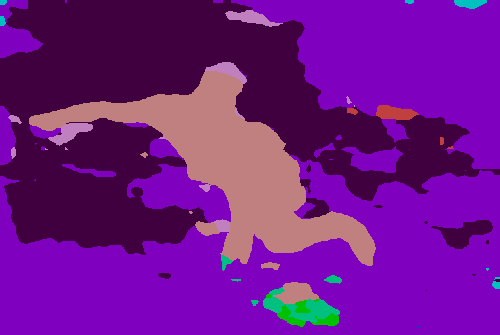}&\includegraphics[width=1.6cm]{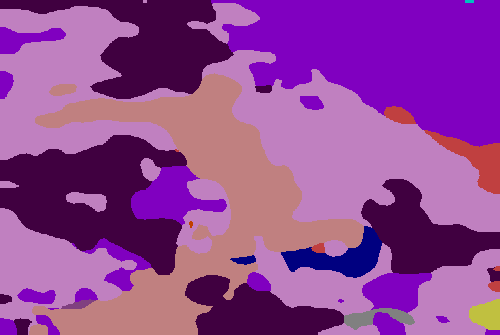}&\includegraphics[width=1.6cm]{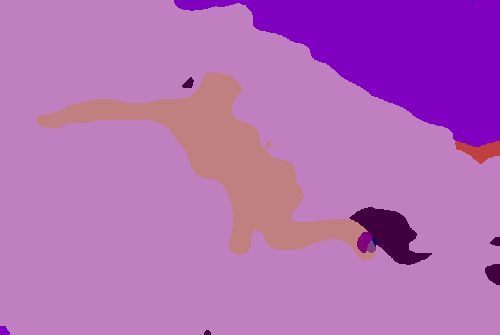}\\
		\tiny{Input} & \tiny{Groundtruth} & \tiny{FCN-8s} & \tiny{plain DAG-RNN} & \tiny{DD-RNNs} \\
	\end{tabular}
	\caption{Qualitative labeling results with VGG-16 on the PASCAL Context~\cite{mottaghi2014role}. Best viewed in color.}
	\label{fig:fig5}
\end{figure}
images, which are crucial for inferring ambiguous pixels. The method in~\cite{shuai2017scene} employs RNNs to capture contextual dependencies among image units for scene labeling and shows outstanding performance with mIoU of 42.6\% with VGG-16. Moreover, they use CRF to improve the result to 43.7\%. Different from~\cite{shuai2017scene}, we propose DD-RNNs to capture richer dependencies. Without any class weighting strategy and post-processing, our DD-RNNs with VGG-16 obtain the mIoU of 44.9\%, which outperforms the method in~\cite{shuai2017scene} by 1.2\%, showing the advantage of DD-RNNs. With deeper ResNet-101, we achieve the mIoU of 49.3\%, outperforming the state-of-the-art RefineNet~\cite{lin2016refinenet}.

Fig.~\ref{fig:fig5} shows qualitative results obtained with VGG-16 on PASCAL Context~\cite{mottaghi2014role}. Without considering long-range contextual dependencies in images, FCN-8s~\cite{long2015fully} is prone to cause misclassification (see the third column in Fig.~\ref{fig:fig5}). Our baseline can help alleviate this situation using RNNs to capture dependencies in images. For example, in the first two rows in Fig.~\ref{fig:fig5}, the `water' can be correctly recognized with the dependencies from `boat'. However, the plain RNNs fail in more complex scenes (see the last three rows in Fig.~\ref{fig:fig5}). For example, in the fourth row in Fig.~\ref{fig:fig5}, most of `road' pixels are mistakenly classified into `ground' pixels without full use of dependencies from `bus'. By contrast, the proposed DD-RNNs are capable of recognizing most of `road' pixels by taking advantages of richer dependencies from `bus' in images.

\subsection{Comparison results on MIT ADE20K}

\begin{table}[htbp]
	\centering
	\caption{Quantitative comparisons on MIT ADE20K (validation set)~\cite{zhou2017scene}.}
	\begin{tabular}{r|ccc}
		\hline
		Algorithm & Backbone & mIoU (\%) \\
		\hline
		\hline
		SegNet~\cite{badrinarayanan2015segnet}  & VGG-16        & 21.6 \\
		FCN-8s~\cite{long2015fully}  & VGG-16      & 29.4 \\
		DilatedNet~\cite{yu2015multi} & VGG-16       & 32.3 \\
		Cascade-SegNet~\cite{zhou2017scene} & VGG-16      & 27.5 \\
		Cascade-Dilated~\cite{zhou2017scene} & VGG-16       & 34.9 \\
		GCE~\cite{hung2017scene} &ResNet-101        & 38.4 \\
		RefineNet~\cite{lin2016refinenet} &ResNet-101        & 40.2 \\
		\hline
		\hline
		DD-RNNs   & VGG-16     &  35.7 \\
		DD-RNNs  & ResNet-101       & {\bf 40.9} \\
		\hline
	\end{tabular}%
	\label{tab:tab2}%
\end{table}%

Table \ref{tab:tab2} summarizes the quantitative results and comparisons to other algorithms. The FCN-8s~\cite{long2015fully} achieves the mIoU of 29.4\%. To incorporate multi-scale contexts,~\cite{yu2015multi} proposes the dilated convolution and improves the mIoU to 32.3\%. To same end, Hung {\it et al.}~\cite{hung2017scene} embed global context into CNNs to obtain improvements, and improve the performance to 38.4\% with ResNet-101. Though the aforementioned methods take the global context of image into account, they ignore long-range contextual dependencies in images. In this work, we employ DD-RNNs to model this dependency information for scene labeling. In specific, we obtain the mIoU of 35.7\% with VGG-16, and achieve better performance with mIoU of 40.9\% when using ResNet-101 as backbone.

\subsection{Comparison results on Cityscapes}

Table~\ref{tab:city} summarizes the quantitative comparison results with state-of-the-art approaches on Cityscapes~\cite{cordts2016cityscapes}. Since the resolution of image is too large, we divide each image into multiple patches. After obtaining the parsing result of each patch, we combine them to derive the labeling of original image. Among the compared algorithms, FCN-8s~\cite{long2015fully} achieves the mIoU of 65.3\%. Liu {\it et al.}~\cite{liu2015semantic} adopt Markov Random Field (MRF) to model high-order CNNs and obtain a mIoU of 66.8\%. The approach of~\cite{lin2016efficient} utilizes CRF to capture contextual information for scene parsing and improves the mIoU to 71.6\%. DeepLabv2~\cite{chen2016deeplab1} combines both CRF and atrous convolution to incorporate more contexts and achieves a mIoU of 70.4\%. In this work, we propose dense RNNs to capture richer dependencies from the whole image for each image unit. With the ResNet backbone, we achieve the mIoU of 78.2\%, outperforming other context aggregation methods.

\begin{table}[htbp]
	\centering
	\caption{Quantitative comparisons on Cityscapes (test set)~\cite{cordts2016cityscapes}.}
	\begin{tabular}{r|ccc}
		\hline
		Algorithm & Backbone & mIoU (\%) \\
		\hline
		\hline
		SegNet~\cite{badrinarayanan2015segnet} &VGG-16 & 57.0 \\
		FCN-8s~\cite{long2015fully}    &VGG-16        &   65.3 \\
		DPN~\cite{liu2015semantic}    & VGG-16        &66.8 \\
		LRR-4x~\cite{ghiasi2016laplacian}  & VGG-16        & 71.8 \\
		CNN-CRF~\cite{lin2016efficient}  & VGG-16      & 71.6 \\
		DilatedNet~\cite{yu2015multi}   & VGG-16    & 67.1 \\
		DeepLab v2-CRF~\cite{chen2016deeplab1}  & ResNet-101 & 70.4 \\
		LC~\cite{li2017not} & ResNet-101 & 71.1 \\
		RefineNet~\cite{lin2016refinenet} & ResNet-101 & 73.6 \\
		PEARL~\cite{jin2017video} & ResNet-101 & 74.9 \\
		SAC~\cite{zhang2017scale} & ResNet-101 & 78.1 \\
		\hline
		\hline
		DD-RNNs  & VGG-16       & 72.3 \\
		DD-RNNs  & ResNet-101       & {\bf 78.2} \\
		\hline
	\end{tabular}%
	\label{tab:city}%
\end{table}%

\begin{table}[htbp]
	\centering
	\caption{Analysis of mIoU (\%) with and without attention model in DD-RNNs using VGG-16.}
	\begin{tabular}{lcc}
		\hline
		& \multicolumn{1}{l}{\tabincell{c}{DD-RNNs w/o \\ attention model}} & \multicolumn{1}{l}{\tabincell{c}{DD-RNNs w/ \\ attention model}} \\
		\hline
		PASCAL Context & 44.3  & {\bf 44.9} \\
		MIT ADE20K & 34.5  & {\bf 35.7} \\
		Cityscapes & 72.0  & {\bf 72.3} \\
		\hline
	\end{tabular}%
	\label{tab:tab4}%
\end{table}%

\begin{figure}[htbp]
	\centering
	\begin{tabular}{@{}C{2.75cm}@{}C{2.75cm}@{}C{2.75cm}@{}}
		\includegraphics[width=2.7cm]{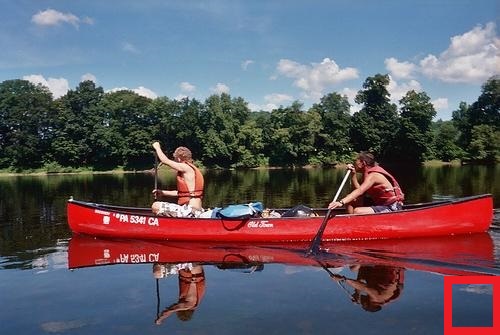}&\includegraphics[width=2.7cm]{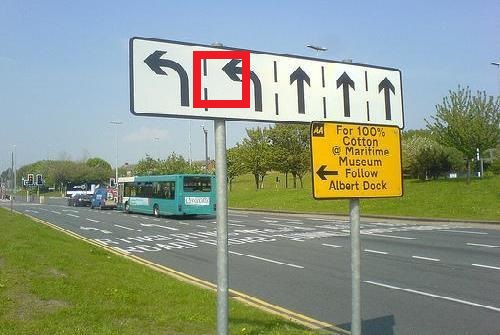}&\includegraphics[width=2.7cm]{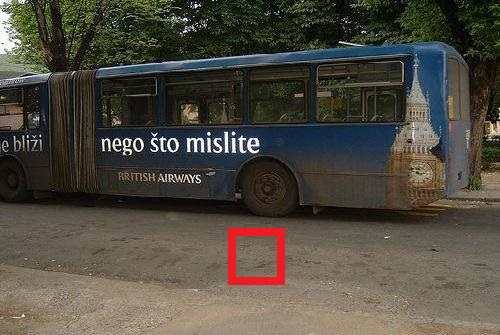}\\
		\includegraphics[width=2.7cm]{res/gt/2008_000834_resize.png}&\includegraphics[width=2.7cm]{res/gt/2010_005123_resize.png}&\includegraphics[width=2.7cm]{res/gt/2010_005756_resize.png}\\
		\includegraphics[width=2.7cm]{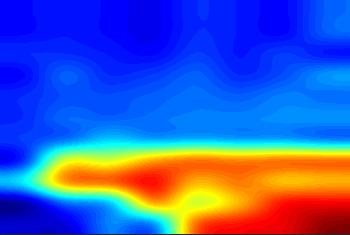}&\includegraphics[width=2.7cm]{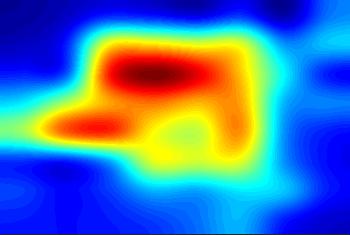}&\includegraphics[width=2.7cm]{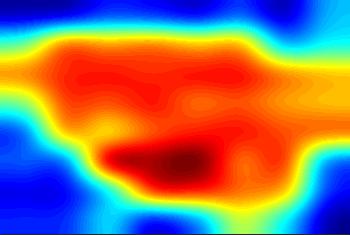}\\
	\end{tabular}
	\caption{Visualization of the learned attentional weight map for a specific region (marked in red rectangle in the first row). First row: input image. Second row: groundtruth. Third row: attentional weight map. Best viewed in color.}
	\label{fig:att}
\end{figure}

\subsection{Ablation study on attention model}

In this paper, we propose the DD-RNNs to model richer dependencies in images, which significantly enhances discriminability for each image unit. However, different dependencies are not always equally helpful. To activate relevant and restrain irrelevant dependencies, we introduce an attention model into DD-RNNs. To demonstrate the effectiveness of attention model, we conduct experiment by removing attention model from DD-RNNs. Note that in these two groups of experiments, the only difference is the attention model, while other settings (e.g., parameters for all other layers) are exactly the same. Table \ref{tab:tab4} summarizes the results on three benchmarks with VGG-16, and shows that the attention model helps to further improve performance.

\begin{table}[htbp]
	\centering
	\caption{Analysis of computation complexity and accuracy of DD-RNNs on PASCAL Context~\cite{mottaghi2014role} dataset.}
	\begin{tabular}{r|C{1.6cm}@{}C{1.55cm}@{}}
		\hline
		Algorithm  & Inference & mIoU (\%)  \\
		\hline
		\hline
		CFM~\cite{dai2015convolutional}         & 0.57 s     & 34.4 \\
		CAMN~\cite{abdulnabi2017episodic}        & 0.27 s    & 41.2 \\
		FCN-8s~\cite{long2015fully}      & 0.32 s    & 38.2 \\
		ParseNet~\cite{liu2015parsenet}      & 0.25 s     & 40.4 \\
		CRF-RNN~\cite{zheng2015conditional}      & 0.70 s     & 39.3 \\
		DeepLab~\cite{chen2016deeplab}     & 0.40 s     & 37.6 \\
		\hline
		\hline
		Baseline FCN (VGG-16)       & 0.17 s    &  35.6 \\
		Baseline DAG-RNN (VGG-16)     & 0.21 s    &  41.3 \\
		DD-RNNs (VGG-16)     & 0.28 s    &  44.9 \\
		DD-RNNs (ResNet-101)     & 0.36 s    &  49.3 \\
		\hline
	\end{tabular}
	\label{tab:tab5}
\end{table}

In order to better understand the attention model, we show the learned attentional weight map for a specific region as shown in Fig.~\ref{fig:att}. From Fig.~\ref{fig:att}, we can see that relevant dependencies are enhanced while irrelevant information are restrained. For example in the first column, the most helpful contextual dependencies for the `water' region come from its surrounding and the `boat' instead of `tree' or `sky', and our attention model learns to pay more importance to the relevant dependencies (i.e., surrounding region and `boat') in the weight map. In the second column in Fig.~\ref{fig:att}, to recognize `sign' region, the useful information comes from surrounding and the `bus', our attention model highlights these regions. In the third column, we can see that our model pays more attention to the relevant `bus' dependencies to correctly recognize the `road' region.

\subsection{Study on model complexity}

As a practical application, both efficiency and accuracy are crucial for scene labeling. To better analyze the proposed approach, we demonstrate the inference time of one forward pass and accuracy on the PASCAL Context~\cite{mottaghi2014role}.

Table~\ref{tab:tab5} reports the efficiency and accuracy of our baseline and other scene labeling algorithms. Compared to its baseline FCN (VGG-16), our algorithm DD-RNNs (VGG-16) obtains mIoU gain of 9.3\% while the inference time only increase by 0.11s, showing the advantage of our DD-RNNs module. Moreover, when replacing the VGG-16 with ResNet-101 as our backbone, the mIoU is further improved to 49.3\%. In comparison with approaches including CAMN~\cite{abdulnabi2017episodic}, FCN-8s~\cite{long2015fully}, CRF-RNN~\cite{zheng2015conditional} and DeepLab~\cite{chen2016deeplab}, our method runs efficiently while achieving better accuracy.

\section{Conclusion}
\label{sec_con}

This paper proposes dense RNNs for scene labeling. Unlike existing methods exploring limited dependencies, our DAG-structured dense RNNs (DD-RNNs) exploit abundant contextual dependencies through dense connections in an image, which better improves the discriminative power of image units. In addition, considering that different dependencies are not always equally helpful to recognize each image unit, we propose an attention model to assign more importance to relevant dependencies. Integrating with CNNs, we develop an end-to-end labeling system. Extensive experiments on PASCAL Context, MIT ADE20K and Cityscapes demonstrate that our DD-RNNs significantly improve the baselines and outperform other state-of-the-art algorithms, evidencing the effectiveness of proposed dense RNNs.

\vspace*{0.5em}
\noindent\textbf{Acknowledgements.} This work is supported in part by US NSF Grants 1350521, 1407156, and 1814745.

{\small
\bibliographystyle{ieee}
\bibliography{egbib}
}

\end{document}